\documentclass[11pt]{article}
\title{}
\author{}
\usepackage{graphicx}
\usepackage{float}

\usepackage[dvipsnames]{xcolor}
\definecolor{cblue}{RGB}{8, 85, 153}
\usepackage[citecolor=cblue, linkcolor=cblue, colorlinks=true, urlcolor=cblue]{hyperref}
\definecolor{mygray}{gray}{0.5}
\definecolor{darkblue}{RGB}{1, 43, 112}

\definecolor{cgreen}{RGB}{8, 153, 83}

\usepackage{amsmath}
\usepackage{amssymb}

\usepackage[fencedCode,inlineFootnotes,citations,definitionLists,hashEnumerators,smartEllipses,hybrid,underscores=false]{markdown} 
\usepackage{placeins} 

\usepackage[hyperpageref]{backref}
\renewcommand*{\backrefalt}[4]{%
    \ifcase #1 \footnotesize{(Not cited.)}%
    \or        \footnotesize{(Cited on page~#2.)}%
    \else      \footnotesize{(Cited on pages~#2.)}%
    \fi}

\usepackage{changepage}

\usepackage{amsthm}

\newtheorem*{assumption*}{\assumptionnumber}
\newtheorem{theorem}{Theorem}
\providecommand{\assumptionnumber}{}


\usepackage[capitalize]{cleveref}
\Crefname{section}{Sec}{Secs.}
\Crefname{figure}{Fig}{Figs.}
\creflabelformat{equation}{#2#1#3}

\usepackage{fontawesome}

\newcommand{\bfp}[1]{(\textbf{#1})}

\usepackage{cite} 

\newcommand{\inner}[2]{\langle{#1},{#2}\rangle} 
\newcommand{\norm}[1]{\lVert{#1}\rVert}

\renewcommand{\O}[1]{\mathcal{O}\left({#1}\right)}
\def\R{\mathbb{R}}
\def\Z{\mathbb{Z}}

\newcommand{\ones}{\mathbf{1}}

\newcommand{\ignore}[1]{}

\newcommand{\loss}{\mathcal{L}}

\newcommand{\Wset}{\mathcal{W}}
\newcommand{\trim}[1]{\textnormal{TRIM}_{\Psi,f}({#1})}
\newcommand{\attr}[1]{\textit{attr}({#1})}

\newcommand{\method}{AWD}
\usepackage{tabularx}
\usepackage{chngcntr}
\usepackage{amsmath,amssymb,amsfonts,graphicx,amsthm,nicefrac,mathtools,bm}
\usepackage[margin=1.1in]{geometry}
\usepackage{xcolor}
\usepackage{rotating}
\usepackage{hyperref}
\usepackage{url}
\usepackage[english]{babel}
\usepackage[margin=1.1in]{geometry}
\usepackage{bbm}
\usepackage{xspace}
\usepackage{booktabs}

\makeatletter
\newcommand{\specificthanks}[1]{\@fnsymbol{#1}}
\makeatother

\title{Adaptive wavelet distillation from neural networks through interpretations}
\author{Wooseok Ha\thanks{Department of Statistics, UC Berkeley} \and
	Chandan Singh\thanks{Department of Electrical Engineering and Computer Sciences, UC Berkeley} \and
	Francois Lanusse\thanks{AIM, CEA, CNRS;  Universit\'e Paris-Saclay, Universit\'e Paris Diderot, Sorbonne Paris Cit\'e} \and
	Srigokul Upadhyayula\thanks{Department of Molecular \& Cell Biology, Advanced Bioimaging Center, UC Berkeley} \and
	Bin Yu\textsuperscript{\specificthanks{1}}\textsuperscript{\specificthanks{2}}
}

\date{}

\begin{document}

\maketitle

\begin{abstract}
 Recent deep-learning models have achieved impressive prediction performance, but often sacrifice interpretability and computational efficiency.
 Interpretability is crucial in many disciplines, such as science and medicine, where models must be carefully vetted or where interpretation is the goal itself.
 Moreover, interpretable models are concise and often yield computational efficiency.
 Here, we propose adaptive wavelet distillation (AWD), a method which aims to distill information from a trained neural network into a wavelet transform.
 Specifically, AWD penalizes feature attributions of a neural network in the wavelet domain to learn an effective multi-resolution wavelet transform.
 The resulting model is highly predictive, concise, computationally efficient, and has properties (such as a multi-scale structure) which make it easy to interpret.
 In close collaboration with domain experts, we showcase how AWD addresses challenges in two real-world settings: cosmological parameter inference and molecular-partner prediction.
 In both cases, AWD yields a scientifically interpretable and concise model which gives predictive performance better than state-of-the-art neural networks.
 Moreover, AWD identifies predictive features that are scientifically meaningful in the context of respective domains.
 All code and models are released in a full-fledged package available on Github.\footnote{\href{https://github.com/Yu-Group/adaptive-wavelets}{\faGithub\,github.com/Yu-Group/adaptive-wavelets}}
\end{abstract}

\section{Introduction}

Recent advancements in deep learning have led to impressive increases in predictive performance.
However, the inability to interpret deep neural networks (DNNs) has led them to be characterized as black boxes.
It is often critical that models are inherently interpretable~\cite{rudin2019stop, rudin2021interpretable,murdoch2019definitions}, particularly in high-stakes applications such as medicine, biology, and policy-making.
In these cases, interpretations which are relevant to a particular domain/audience~\cite{murdoch2019definitions} can ensure that models behave reasonably, identify when models will make errors, and make the models more amenable to inspection and improvement by domain experts.
Moreover, interpretable models tend to be faster and more computationally efficient than large neural networks.

One promising approach to constructing interpretable models without sacrificing prediction performance is model distillation.
Model distillation~\cite{hinton2015distilling,xu2020computation,guo2020spherical} transfers the knowledge in one model (i.e., the teacher), into another model (i.e., the student), where the student model often has desirable properties, such as being more interpretable than the teacher model.
Recent works have considered distilling a DNN into inherently interpretable models such as a decision tree~\cite{frosst2017distilling,craven1996extracting,li2020tnt} or a global additive model~\cite{tan2018learning}, with some success. Here, we consider distilling a DNN into a learnable wavelet transform, which 
is a powerful tool to describe signals both in time (spatial) and frequency domains that has found numerous successful applications in physical and biomedical sciences.

\begin{figure}[H]
	\centering
	\includegraphics[width=1\textwidth]{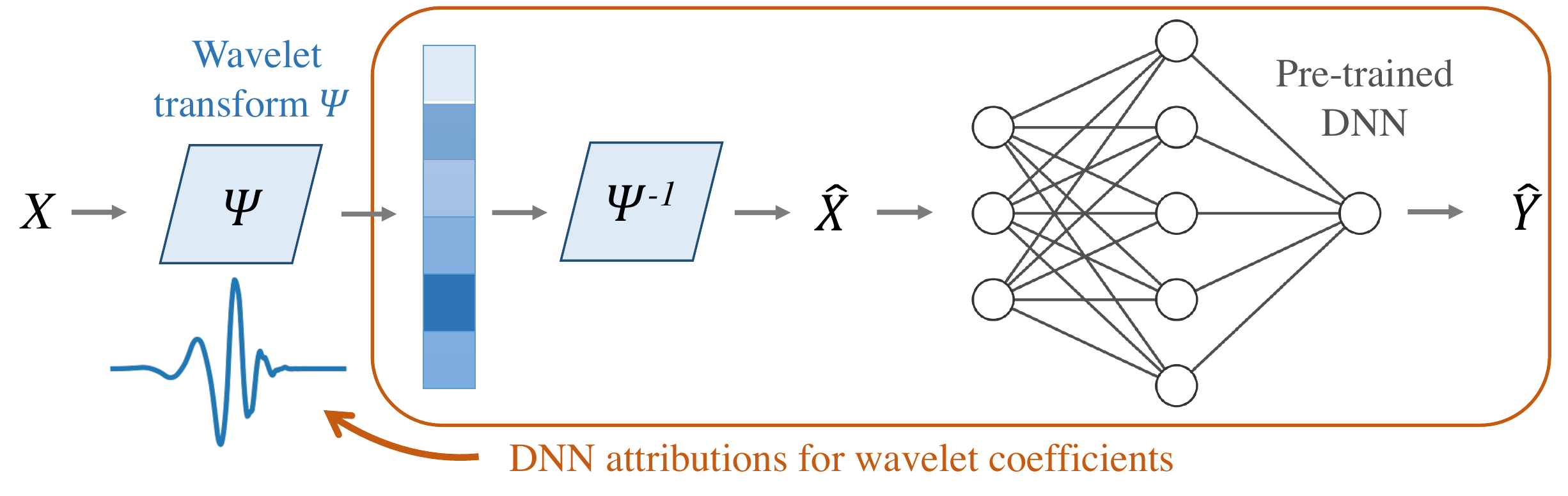}
	\caption{Adaptive wavelet distillation uses attributions from a trained DNN to improve its wavelet transform, while satisfying constraints for reconstruction error and wavelet constraints.}
	\label{fig:intro}
\end{figure}

Wavelets have many properties amenable to interpretation: they can form an orthogonal basis, identify a sparse representation of a signal, and tile different frequencies and spatial locations (and sometimes rotations), allowing for multiresolution analysis. 
Most previous work has focused on hand-designed wavelets for different scenarios rather than wavelets which adapt to given data.
Recent work has explored wavelets which adapt to an input data distribution, under the name optimized wavelets or adaptive wavelets~\cite{recoskie2018learning,sallee2002learning,lin2003gearbox,brechet2007compression,hereford2003image,recoskie2018learningthesis,jawali2019learning,tai2016multiscale}.
Moreover, some work has used wavelets as part of the underlying structure of a neural network, as in wavelet networks~\cite{zhang1992wavelet}, wavelet neural networks~\cite{zhang1995wavelet,DHEEBA201445}, or the scattering transform~\cite{mallat2012group,bruna2013invariant}. 
However, none of them utilize wavelets for model interpretability. 

\cref{fig:intro} outlines \underline{A}daptive \underline{W}avelet \underline{D}istillation (\method), our approach for distilling a wavelet transform from a trained DNN. A key novelty of AWD is that it uses \textit{attributions from a trained DNN} to improve the learned wavelets;\footnote{By attributions, we mean feature importance scores given input data and a pre-trained DNN.} this incorporates information not just about the input signals, as is done in previous work, but also about the target variable and the inductive biases present in the DNN.\footnote{Though we focus on DNNs, AWD works for any black-box models for which we can attain attributions.}

This paper deviates significantly from a typical NeurIPS paper. While there has been an explosion of work in ``interpretable machine learning''~\cite{molnar2019interpretable}, there has been very limited development and grounding of these methods \textit{in the context of a particular problem and audience}. This has led to much confusion about how to develop and evaluate interpretation methods~\cite{adebayo2018sanity,doshi2017roadmap}; in fact, a major part of the issue is that interpretability cannot be properly defined without the context of a particular problem and audience~\cite{murdoch2019definitions}.
As interpretability and scientific machine learning enter a new era, researchers must ground themselves in real-world problems and work closely with domain experts.

This paper focuses on scientific machine learning---providing insight for a particular scientific audience into a chosen scientific problem--- and from its outset, was designed to solve a particularly challenging cosmology problem in close collaboration with cosmologists. We showcase how AWD can inform relevant features in a fundamental problem in cosmology: inferring cosmological parameters from weak gravitational lensing convergence maps.\footnote{For the purpose of this work, we work with simulated lensing maps.} In this case, AWD identifies high-intensity peaks in the convergence maps and yields an easily interpretable model which outperforms state-of-the-art neural networks in terms of prediction performance.
We next find that AWD successfully provides prediction improvements in another scientific application (now in collaboration with cell-biology experts): molecular-partner prediction. 
In this case, AWD allows us to vet that the model’s use of clathrin corresponds to our domain knowledge about how clathrin must build up slowly then fall in order to predict a successful event.
In both cases, the wavelet models from AWD further extract compressed representations of the input in comparison to the standard wavelet model while concisely explaining model behavior.
We hope that the depth and grounding of the scientific problems in this work can spur further interpretability research in real-world problems, where interpretability can be evaluated by and enrich domain knowledge, beyond benchmark data contexts such as MNIST~\cite{lecun1998mnist} where the need for interpretability is less cogent.



\section{Background on wavelet transform and TRIM}
\label{sec:background}

\subsection{Wavelet transform}
\label{sec:wave_transform}

Wavelets are a class of functions that are localized both in the time and frequency domains. In the classical setting, each wavelet is a variation of a single wavelet $\psi$, called the \textit{mother wavelet}. A family of discrete wavelets can be created by scaling and translating the mother wavelet in discrete increments:
\begin{equation}\label{eqn:wavelets}
    \left\{\psi_{j, n}(t)=\frac{1}{\sqrt{2^{j}}} \psi\left(\frac{t-2^{j} n}{2^{j}}\right)\right\}_{(j, n) \in \mathbb{Z}^{2}},
\end{equation}
where each wavelet in the family $\psi_{j, n}(t)$ represents a unique scale and translation of $\psi$. With a carefully constructed wavelet $\psi$ (see \cref{subsec:orthonormal_basis}), the family of wavelets~\eqref{eqn:wavelets} forms an orthonormal basis of $L^2(\R)$. Namely, any signal $x\in L_2(\R)$ can be decomposed into 
\begin{equation}\label{eqn:wavelet_decomp0}
    x=\sum_{n} \sum_{j} d_{j}[n] \psi_{j, n},
\end{equation}
where the wavelet (or detail) coefficients $d_{j}[n]$ at scale $2^j$ are computed by taking the inner product with the basis functions, $d_{j}[n] = \inner{x}{\psi_{j,n}} = \int x(t)\psi_{j,n}(t)dt.$ The decomposition~\eqref{eqn:wavelet_decomp0} requires an infinite number of scalings to calculate the discrete wavelet transform. To make this decomposition computable, the {\em scaling function} $\phi$ is introduced so that 
\begin{equation}\label{eq:wavelet_decomp}
    x=\sum_{n} a_{J}[n] \phi_{J, n}+\sum_{n} \sum_{j}^{J} d_{j}[n] \psi_{j, n},
\end{equation}
where $\phi_{J, n}(t)=2^{-J/2}\phi(2^{-J}t-n)$ represent different translations of $\phi$ at scale $2^J$ and $a_{J}[n]= \inner{x}{\phi_{J,n}}$ are the corresponding approximation coefficients. Conceptually, the $\phi_{J, n}$ form an orthogonal basis of functions that are smoother at the given scale $2^J$, and therefore can be used to decompose the smooth residuals not captured by the wavelets~\cite{mallat2008wavelet}.

A fundamental property of the discrete wavelet transform is that the approximation and detail coefficients at scale $2^{j+1}$ can be computed from the approximation coefficients of the previous scale at $2^j$~\cite{mallat1989multiresolution,meyer1992wavelets}. To see this, let us define the two discrete filters, lowpass filter $h$ and highpass filter $g$
\begin{equation}\label{eqn:filters}
    h[n] = \inner{\frac{1}{\sqrt{2}}\phi(t/2)}{\phi(t-n)} \ \ \text{ and } \ \ g[n] = \inner{\frac{1}{\sqrt{2}}\psi(t/2)}{\phi(t-n)}.
\end{equation}
Then the following recursive relations hold between the approximation and detail coefficients at two consecutive resolutions:
\begin{equation}\label{eqn:recursive_analysis}
    \begin{cases}
    a_{j+1}[p] = \sum_n h[n-2p]a_j[n] = a_j\star\bar{h}[2p]; \\
    d_{j+1}[p] = \sum_n g[n-2p]a_j[n] = a_j\star\bar{g}[2p],
    \end{cases}
\end{equation}
where we denote $\bar{h}[n]=h[-n]$ and $\bar{g}[n]=g[-n]$. Conversely, the approximation coefficients at scale $2^j$ can be recovered from the coarser-scale approximation and detail coefficients using
\begin{equation}\label{eqn:recursive_synthesis}
    a_j[p] = \sum_n h[p-2n]a_{j+1}[n] + \sum_n g[p-2n]d_{j+1}[n].
\end{equation}
Together, these recursive relations lead to the filter bank algorithm, the cascade of discrete convolution and downsampling, which can be efficiently implemented in time $\O{\textnormal{Signal length}}$. The discrete wavelet transform can be extended to two dimensions, using a separable (row-column) implementation of 1D wavelet transform along each axis (see~\cref{sec:2d_wave_transform}).

\subsection{Transformation Importance (TRIM)}
\label{sec:TRIM}

The work here requires the ability to compute attributions which identify important features given input data and a trained DNN.
Most work on interpreting DNNs has focused on attributing importance to features in the input space of a model, such as pixels in an image or words in a document~\cite{ribeiro2016should,lundberg2017unified,sundararajan2016gradients,singh2018hierarchical,rieger2019interpretations}.
Instead, here we rely on TRIM (Transformation Importance)~\cite{singh2020transformation}, an approach which attributes importance to features in a transformed domain (here, the wavelet domain) via a straightforward model reparameterization.

Formally, let $f$ be a pre-trained model that we desire to interpret. If $\Psi:\mathcal{X}\to \mathcal{W}$ is a bijective mapping that maps an input $x$ to a new domain $w=\Psi(x)\in\Wset$, TRIM reparameterizes the model as $f'=f\circ \Psi^{-1}$, where $\Psi^{-1}$ denotes the inverse of $\Psi$. In the case that $\Psi$ is not exactly invertible, TRIM adds the residuals to the output of $\Psi^{-1}$, i.e., $f'$ is reparameterized by $f'(w)=f(\Psi^{-1}w +r)$ where $w=\Psi(x)$ and $r= x - \Psi^{-1}(w)$. If $S$ indexes a subset of features in the transformed space indicating which part of the transformed input to interpret, we then define 
\begin{equation}\label{eqn:TRIM}
    \trim{w_S} = \attr{f';w_S},
\end{equation}
where $\attr{;w}$  is an attribution method that is evaluated at $w$ and outputs an importance value, and where $w_S$ denotes the subvector of $w$ indexed by $S$. The choice of the attribution method $\attr{}$ can be any local interpretation technique (e.g. LIME~\cite{ribeiro2016should} or Integrated Gradients (IG)~\cite{sundararajan2016gradients}); here we focus mainly on the saliency map~\cite{simonyan2013deep}, which simply calculates the gradient of the model's output with respect to its transformed input to define feature attribution.  We leave more complex attribution methods such as IG or ACD~\cite{singh2018hierarchical} to future work. 
\section{Adaptive wavelet distillation through interpretations}
\label{sec:methods}

Adaptive wavelet distillation (AWD) aims to learn a wavelet transform which effectively represents the input data as well as captures information about a model trained to predict a response using the input data. Depending on a problem's context, the resulting wavelet model may or not be sufficiently interpretable for use, or may or not provide similar or better prediction performance as the pre-trained model $f$. However, we provide two scientific data problems in section~\ref{sec:results} where we can do both. In fact, the AWD method has been developed in the context of solving the cosmology problem.

We now detail how AWD wavelets can be built upon to form an extremely simple model in various contexts (see~\cref{sec:results}). We first require that the wavelet transform is invertible, allowing for reconstruction of the original data. This ensures that the transform does not lose any information in the input data.
We next assure that the learned wavelet is a valid wavelet: the wavelet function $\psi$  and the corresponding scaling function $\phi$ span a sequence of subspaces satisfying the multiresolution axioms~\cite{mallat1989theory}. Finally, we add the distillation part of AWD.
We calculate the attribution scores of a given model $f$ for each coefficient in the wavelet representation, and try to find a wavelet function $\psi$ that makes these attributions sparse. Intuitively, this ensures that the learned wavelet should find a representation which can concisely explain a model's prediction.
Writing the discrete wavelet transform using the discrete filters $h$ and $g$~(see \cref{eqn:filters}), we now give a final optimization problem for \method:
\begin{equation}
    \label{eqn:objective}
    \underset{h,g}{\text { minimize }}\loss(h,g)= \underbrace{\frac{1}{m}\sum_{i}\norm{x_i - \widehat{x}_i}_{2}^{2}}_{\text {Reconstruction loss }}
    +\underbrace{\frac{1}{m}\sum_i W(h, g, x_i; \lambda)}_{\text {Wavelet loss }}
    +\underbrace{\gamma \sum_{i} \norm{\trim{\Psi x_i}}_1}_{\text {Interpretation loss }},
\end{equation}
where $\Psi$ denotes a wavelet transform operator induced by $\psi$, and $\widehat{x}_i$ denotes the reconstruction of the data point $x_i$. Here $\lambda, \gamma>0$ represent hyperparameters that are tuned by users. The only parameters optimized are the lowpass filter $h$ and the highpass filter $g$. The corresponding scaling and wavelet functions can be obtained from $(h,g)$ via the following mapping~\cite{mallat2008wavelet}: $\widehat{\phi}(w)=\prod_{p=1}^{\infty}\frac{\widehat{h}(2^{-p}w)}{\sqrt{2}}$ and $\widehat{\psi}(w)=\frac{1}{\sqrt{2}} \widehat{g}(w / 2) \widehat{\phi}(w / 2)$, where $\widehat{\phi}$ and $\widehat{\psi}$ represent the Fourier transforms of $\phi$ and $\psi$ respectively.

\paragraph{Wavelet loss} The wavelet loss ensures that the learned filters yield a valid wavelet transform. In contrast to the wavelet constraints used in~\cite{recoskie2018learning}, our formulation introduces additional terms that ensure almost sufficient and necessary conditions on the filters $(h,g)$ to build an orthogonal wavelet basis. Specifically, \cite[Theorem 7.2]{mallat2008wavelet} states the following sufficient conditions on the lowpass filter: if $h$ satisfies 
\begin{equation}\label{eqn:CMF}
    \sum_n h[n] = \sqrt{2} \ \ \text{ and } \ \ |\widehat{h}(w)|^{2}+|\widehat{h}(w+\pi)|^{2}=2 \ \text{ for all $w$},
\end{equation}
as well as some mild conditions, it can generate a scaling function such that the scaled and translated family of the scaling function forms an orthonormal basis of the space of multiresolution approximations of $L^2(\R)$. \cite[Theorem 3]{burrus1997introduction} further shows that the orthogonality of translates of the scaling function implies that the lowpass filter is orthogonal after translates by $2$, i.e.,
\begin{equation}\label{eqn:h_constraints}
    \sum_{n} h[n] h[n-2k] = \begin{cases}1 & \text{if $k=0$} \\ 0  & \text{otherwise} \end{cases}, \ \  \text{ and as a result, $\norm{h}_2=1$}.
\end{equation}
Hence the conditions \eqref{eqn:CMF},~\eqref{eqn:h_constraints} characterize the almost sufficient and necessary conditions on the lowpass filter. Moreoever, \cite[Theorem 7.3]{mallat2008wavelet} shows that the valid highpass filter can be constructed from the lowpass filter: in the time domain, it can be written as
\begin{equation}\label{eqn:g_constraints}
    g[n]=(-1)^{n} h[N-1-n],
\end{equation}
where $N$ is the support size of $h$. Together with~\eqref{eqn:h_constraints}, we can also deduce that the highpass filter has mean zero, i.e., $\sum_n g[n]=0$ which is necessary for the filter $g$. See~\cref{subsec:orthonormal_basis} for further details.

Finally, we want the learned wavelet to provide sparse representations so we add the $\ell_1$ norm penalty on the wavelet coefficients. Combining all these constraints via regularization terms, we define the wavelet loss at the data point $x_i$ as
\begin{multline}\label{eqn:wave_loss}
    W(h,g,x_i;\lambda) = \lambda \norm{\Psi x_i}_1 +  (\sum_n h[n]-\sqrt{2})^2 + (\sum_n g[n])^2 + (\norm{h}_2^2-1)^2 \nonumber\\
    + \sum_{w}(|\widehat{h}(w)|^{2}+|\widehat{h}(w+\pi)|^{2} - 2)^2 + \sum_k(\sum_{n} h[n] h[n-2k]-\ones_{k=0})^2,
\end{multline}
where $g$ is set as in~\eqref{eqn:g_constraints} and $\lambda>0$ controls strength of the sparsity of the wavelet representations. We enforce the penalty $(|\widehat{h}(w)|^{2}+|\widehat{h}(w+\pi)|^{2} - 2)^2$, only at the discrete values of $w\in\{\frac{2\pi k}{N},k=1,\ldots,N\}$ through the  discrete Fourier transform. Notice that the wavelet loss does not introduce any additional hyperparameters besides $\lambda$. In fact, we empirically observe that the sum of penalty terms, except the sparsity penalty, remains very close to zero as long as the filters $(h,g)$ are initialized using known wavelet filters and the interpretation loss is not enforced too strongly.

\paragraph{Interpretation loss}

The interpretation loss enables the distillation of knowledge from the pre-trained model $f$ into the wavelet model. It ensures that attributions in the space of wavelet coefficients $\Psi x_i$ are sparse, where the attributions of wavelet coefficients is calculated by TRIM~\cite{singh2020transformation}, as described in \cref{sec:TRIM}. This forces the wavelet transform to produce representations that concisely explain the model's predictions at different scales and locations.  The hyperparameter $\gamma$ controls the overall contribution of the interpretation loss; large values of $\gamma$ can result in large numerical differences from satisfying the conditions of the mathematical wavelet filters. To our knowledge, this is the first method which uses interpretations from a pre-trained model to improve a wavelet representation. This enables the wavelets to not only adapt to the distribution of the inputs, but also gain information about the predicted outputs through the lens of the model $f$.

\section{AWD improves interpretability, prediction performance, and compression in two scientific problems and in simulations}
\label{sec:results}

\cref{fig:sim} shows a visual schematic of the distillation and prediction setup for one synthetic and two scientific data problems in this section, whose details will be discussed in the following subsections.\footnote{In all experiments, the wavelet function is computed from the corresponding lowpass filter using the \textit{PyWavelets} package~\cite{lee2019pywavelets} and building on the \textit{Pytorch Wavelets}~\cite[Chapter 3]{cotter2020uses} package.}

\begin{figure}[t]
    \centering
    \includegraphics[width=\textwidth]{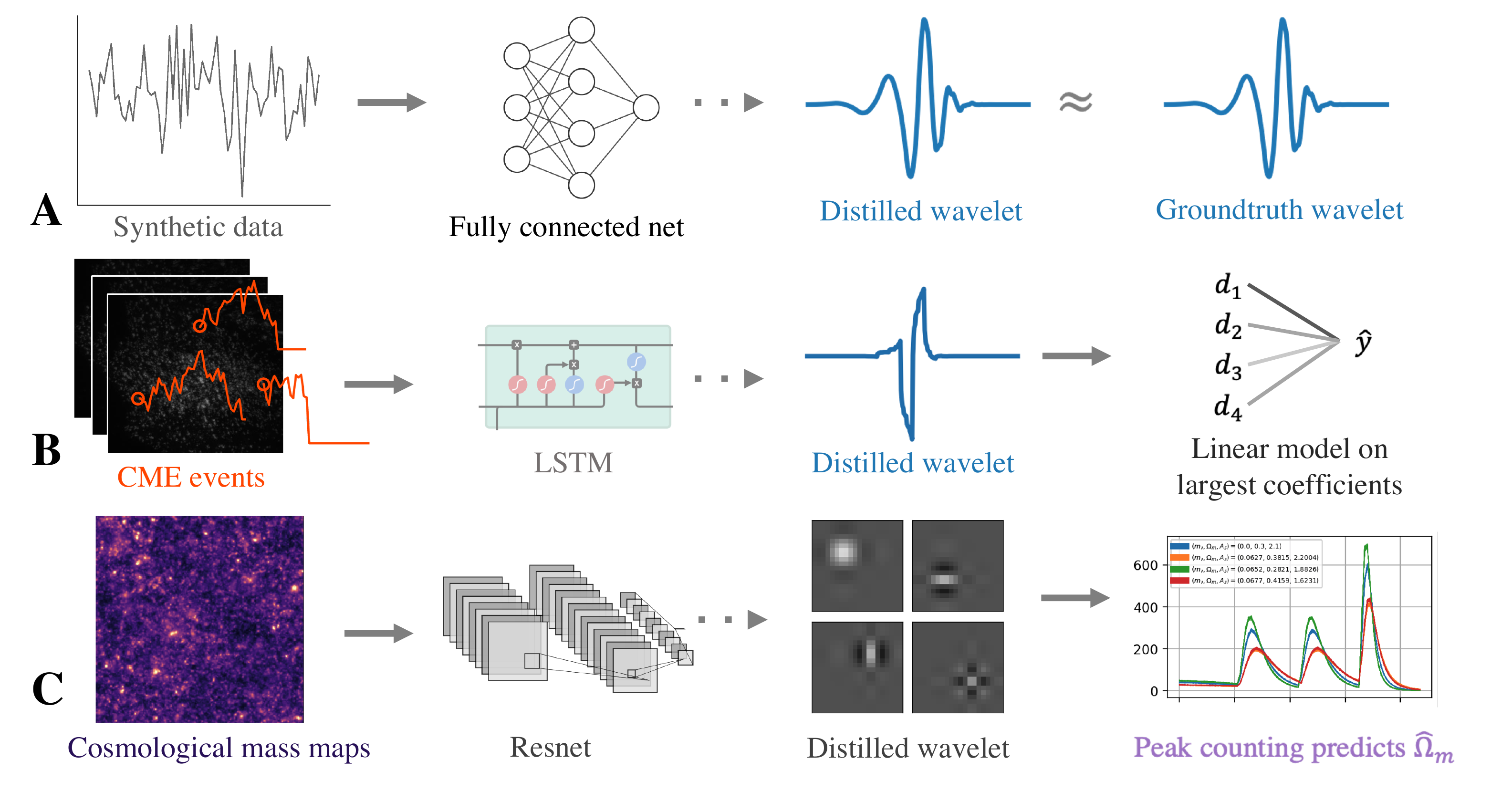}
    \caption{Distillation and prediction setup for the three scenarios in \cref{sec:results}. \bfp{A} In synthetic simulations, AWD is able to recover groundtruth wavelet (DB5) that are linked to a response variable~(\cref{subsec:results_sim}). \bfp{B} Wavelets distilled by AWD from an LSTM trained to predict molecular partners capture biologically meaningful properties of a large build up in clathrin fluorescence followed by a sharp drop and enable prediction using only a few key coefficients~(\cref{subsec:results_molecular}). \bfp{C} AWD finds wavelets that are efficient at capturing cosmological information in weak lensing convergence maps and can improve state-of-the-art performance of cosmological parameter inference using an AWD-based simple peak-counting algorithm~(\cref{subsec:results_cosmo}).}
    \label{fig:modeling}
\end{figure}

\subsection{Synthetic data}
\label{subsec:results_sim}

We begin our evaluation using simulations to verify whether AWD can recover groundtruth wavelets from noisy data. In these simulations, the inputs $x_i$ are generated i.i.d. from a standard Gaussian distribution $\mathcal{N}(0,1)$. To generate the response variable, the inputs are transformed into the wavelet domain using Daubechies (DB) 5 wavelets~\cite{daubechies1988orthonormal}, and the response is generated from a linear model $y_i=\inner{\Psi x_i}{\beta}+\epsilon_i$, where the true regression coefficients are $2$ for a few selected locations at a particular scale and $0$ otherwise; the noise $\epsilon_i$ is generated i.i.d. from a Gaussian distribution $\mathcal{N}(0,0.1^2)$. Then, a 3-layer fully connected neural network with ReLU activations is trained on the pairs of $x_i, y_i$ to accurately predict this response. Note that for any non-singular matrix $A$, the mapping $x\mapsto\inner{A^{-1}\Psi x}{A^\top\beta}$ predicts the response equally well, but the representations in the groundtruth wavelet explain the model's prediction most concisely. The challenge is then to accurately distill the groundtruth wavelet (DB5) from this DNN. This task is fairly difficult: AWD must not only select which scale and locations are important, it must also precisely match the shape of $h$ and $g$ to the groundtruth wavelet.

\begin{figure}[t]
    \centering
    \includegraphics[width=\textwidth]{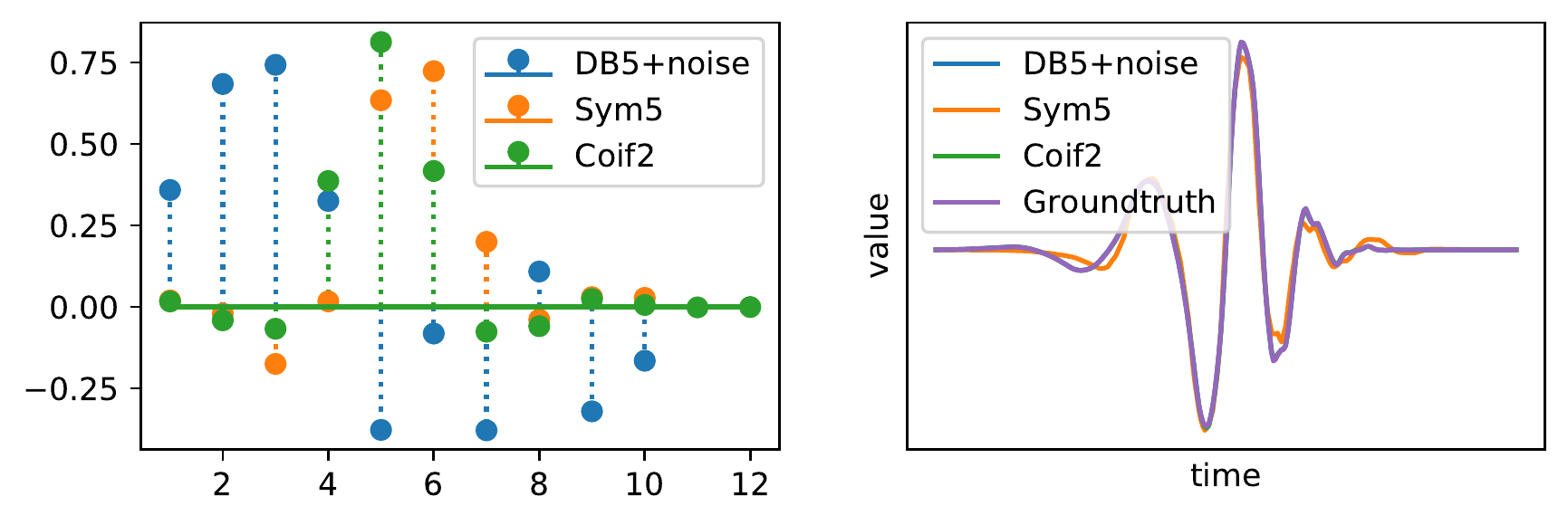}
    \caption{AWD accurately identifies the groundtruth important wavelet in simulated data. \bfp{A} Plots of the
    initial lowpass filters. \bfp{B} Final wavelets extracted by AWD.}
    \label{fig:sim}
\end{figure}

\cref{fig:sim} shows the performance of AWD on this task. We initialize the AWD lowpass filter to different known lowpass filters corresponding to DB5 (and add noise), Symlet 5, and Coiflet 2 (shown in~\cref{fig:sim}\bfp{A}) and then optimize the objective given in \cref{eqn:objective}. In order to recover the groundtruth, we select hyperparameters $\lambda$ and $\gamma$ that minimize the distance to the groundtruth wavelet $\psi^\star$.
Distance is measured via $\textnormal{d}(\psi,\psi^\star)=\min\{\min_k\norm{\psi^k-\psi^\star}_2, \min_k\norm{\widetilde{\psi}^k-\psi^\star}_2\}$, where $\psi^k$ is the wavelet $\psi$ circular shifted by $k$ and $\widetilde{\psi}$ is the wavelet $\psi$ flipped in the left/right direction. That is, $\textnormal{d}$ calculates the minimum $\ell_2$ distance between two wavelets under circular shifts and left/right flip. When the two wavelets have different size of support, the shorter wavelet is zero-padded to the length of the longer~\cite{recoskie2018learning}. \cref{fig:sim}\bfp{B} shows that for each different initialization, we find that the distilled wavelet gets considerably closer to the groundtruth wavelet. In particular, the results for DB5+noise and Coiflet 2 are nearly identical to the groundtruth and cannot be distinguished in the plot.
This is particularly impressive since the support size of Coiflet 2 differs from that of the groundtruth wavelet, making the task more difficult.
Overall, these results demonstrate the ability of AWD to distill key information out of a pre-trained neural network. 


\subsection{Molecular partner-prediction for a central process in cell biology}
\label{subsec:results_molecular}

We now turn our attention to a crucial question in cell biology related to the internalization of macromolecules via clathrin-mediated endocytosis (CME)~\cite{kirchhausen2014molecular}.
CME is the primary pathway for entry into the cell, making it essential to eukaryotic life~\cite{mcmahon2011molecular}.
CME is an orchestra consisting of hundreds of different protein dynamics, prompting a line of research aiming to better understand this process \cite{kaksonen2018mechanisms}.
Crucial to understanding CME is the ability to readily distinguish whether or not the recruitment of certain molecules will allow for endocytosis, i.e., successfully transporting an object into a cell.
Previous approaches have largely relied on the presence of a specific scission/uncoating marker during imaging~\cite{he2020dynamics,wang2020dasc}. Alternatively, previous works use domain knowledge to hand-engineer features based on the lifetime of an event or thresholds on the recruited amplitude of the clathrin molecule \cite{aguet2013advances, kadlecova2017regulation}.

Here, we aim to identify successful CME events with a learning approach, obviating the need for an auxiliary marker or hand-engineered features.
We use a recently published dataset~\cite{he2020dynamics} which tags two molecules: clathrin light chain A, which is used as the predictor variable, and auxilin 1, the target variable.
In this context, clathrin is used to track the progress of an event, (as recruitment of clathrin molecules usually precedes scission) and recruitment of auxilin molecules follows only when endocytosis successfully occurs (to facilitate disassembly of the clathrin-coated vesicle). See data details in \cref{sec:bio_data_details}.
Time-series of fluorescence amplitudes (see \cref{fig:modeling}B) are extracted from raw cell videos for clathrin~\cite{aguet2013advances} and used to predict the mean amplitude of the auxilin signal, an indicator of whether an event was successful or not.
The dataset is randomly split into a training set of 2,936 data units of dimension 40 and a test set of 1,005 data units.
This is a challenging problem where deep learning has recently been shown to outperform classical methods. We train a DNN (an LSTM~\cite{hochreiter1997long}, see architecture in the Supplement) to predict the auxilin response from the clathrin signal.
The model predicts well, but has extremely poor interpretability and computational cost, so we aim here to distill it into a wavelet model through AWD.

\begin{figure}[H]
    \centering
    \includegraphics[width=\textwidth]{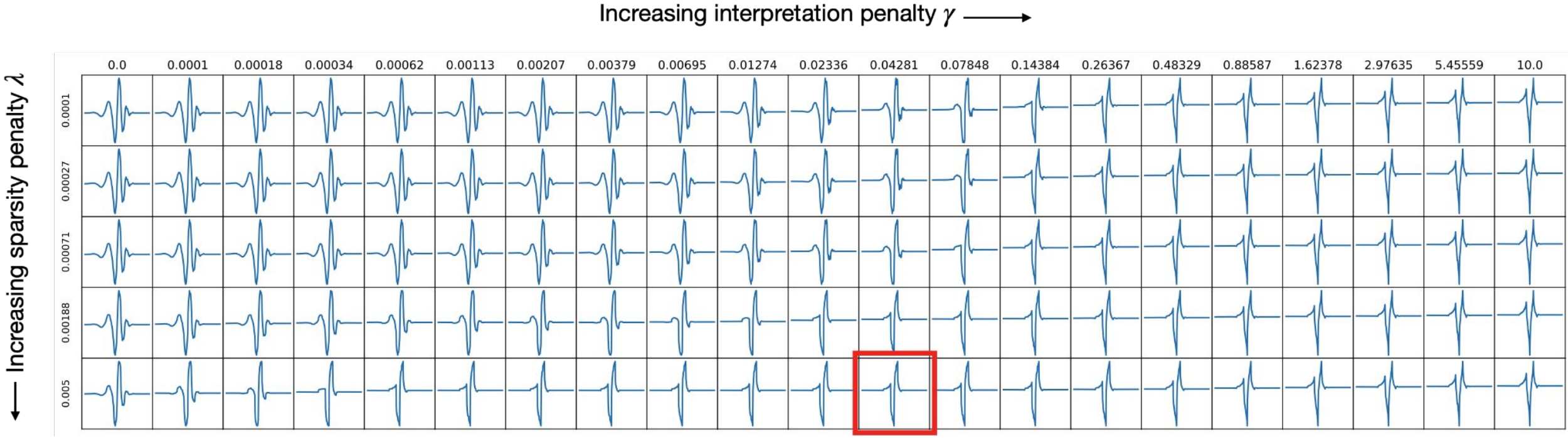}
    \caption{Varying sparsity and interpretation penalty yields different valid wavelets. Wavelet highlighted in red is selected by cross-validation and yields the best prediction performance.}
    \label{fig:vary_wavelet}
\end{figure}

\cref{fig:vary_wavelet} shows qualitatively how the learned wavelet function $\psi$ changes as a function of the interpretation penalty $\gamma$ (increasing to the right) and the sparsity penalty $\lambda$ (increasing downwards). 
In the initial stage of training, we initialize the lowpass filter to correspond to the Daubechies (DB) 5 wavelet. Different combinations of the penalties lead to vastly different learned wavelets, though they all tend to reveal edge-detecting characteristics for a reasonable range of hyperparameter values. 

We now test the distilled wavelets for their predictive power. To create an extremely transparent model, we extract only the maximum $6$ wavelet coefficients at each scale.
By taking the maximum coefficients, these features are expected to be invariant to the specific locations where a CME event occurs in the input data.
This results in a final model with $30$ coefficients ($6$ wavelet coefficients at $5$ scales).
These wavelet coefficients are used to train a linear model, and the best hyperparameters are selected via cross-validation on the training set.
\cref{fig:modeling} shows the best learned wavelet (for one particular run) extracted by AWD corresponding to the setting of hyperparameters $\lambda=0.005$ and $\gamma=0.043$. \cref{tab:auxilin_results} compares the results for \method~to the original LSTM and the initialized, non-adaptive DB5 wavelet model, where the performance is measured via a standard $R^2$ score, a proportion of variance in the response that is explained by the model.
The AWD model not only closes the gap between the standard wavelet model (DB5) and the neural network, it considerably improves the LSTM's performance (a 10\% increase of $R^2$ score). Moreover, we calculate the compression rates of the AWD wavelet and DB5---these rates measure the proportion of wavelet coefficients in the test set, in which the magnitude and the attributions are both above $10^{-3}$. The AWD wavelet exhibits much better compression than DB5 (an 18\% reduction), showing the ability of AWD to simultaneously provide sparse representations and explain the LSTM's predictions concisely. The AWD model also dramatically decreases the computation time at test time, a more than $200$-fold reudction when compared to LSTM.

In addition to improving prediction accuracy, AWD enables
domain experts to vet their experimental pipelines by making them more transparent.
By inspecting the learned wavelet, AWD allows for checking what clathrin signatures signal a successful CME event; it indicates that the distilled wavelet aims to identify a large buildup in clathrin fluorescence (corresponding to the building of a clathrin-coated pit) followed by a sharp drop in clathrin fluorescence (corresponding to the rapid deconstruction of the pit).
This domain knowledge is extracted from the pre-trained LSTM model by \method~using only the saliency interpretations in the wavelet space.

\begin{table}[H]
	\centering
	\caption{Performance comparisons for different models. AWD substantially improves predictive accuracy, compression rate, and computation time on the test set. A higher $R^2$ score, and lower compression factor, and lower computation time indicate better results. For AWD, values are averaged over $5$ different random seeds.}
	
	\begin{tabular}{lrrr}
		\toprule
		{} &  \textbf{AWD (Ours)} &  Standard Wavelet (DB5) &   LSTM \\
		\midrule
		Regression ($R^2$ score)   &     \textbf{0.262 (0.001)} &          0.197  &      0.237 \\
		Compression factor            &       \textbf{0.574 (0.010)}     &     0.704        &     N/A\\
		Computation time        &      \textbf{0.0002s} &    \textbf{0.0002s}  &    0.0449s\\
		\bottomrule
	\end{tabular}

	\label{tab:auxilin_results}
\end{table}

\subsection{Estimating a fundamental parameter surrounding the origin of the universe}
\label{subsec:results_cosmo}

We now focus on a cosmology problem, where AWD helps replace DNNs with a more interpretable alternative.
Specifically, we consider weak gravitational lensing convergence maps, i.e., maps of the mass distribution in the universe integrated up to a certain distance from the observer. In a cosmological experiment (e.g. a galaxy survey), these mass maps are obtained by measuring the distortion of distant galaxies caused by the deflection of light by the mass between the galaxy and the observer~\cite{Bartelmann2001}. These maps contain a wealth of physical information of interest, such as the total matter density in the universe, $\Omega_m$.
Current cosmology research aims to identify the most informative features in these maps for inferring the cosmological parameters such as $\Omega_m$. 
The traditional summary statistic for lensing maps is the power spectrum which is known to be sub-optimal for parameter inference. Tighter parameter constraints can be obtained by including higher-order statistics, such as the bispectrum~\cite{Coulton2019} and peak counts~\cite{LiuPeaks2019}. However, DNN-based inference methods claim to improve on constraints based on these traditional summaries~\cite{Zorilla2019, ribli2019improved, Fluri2019}. 

Here, we aim to improve the predictive power of DNN-based methods while gaining interpretability by distilling a predictive AWD model.
In this context, it is critically important to obtain interpretability, as it provides deeper understanding into what information is most important to infer $\Omega_m$ 
and can be used to design optimal experiments or analysis methods. Moreover, because these models are trained on numerical simulations (realizations of the Universe with different cosmological parameters), it is important to validate that the model uses reasonable features rather than latching on to numerical artifacts in the simulations.
We start by training a model to predict $\Omega_m$ from simulated weak gravitational lensing convergence maps. We train a DNN\footnote{The model's architecture is Resnet 18 \cite{he2016deep}, modified to take only one input channel.} to predict $\Omega_m$ from 100,000 mass maps simulated 
with 10 different sets of cosmological parameter values at the universe origin
from the \texttt{MassiveNuS} simulations \cite{Jia2018} (full simulation details given in \cref{sec:cosmo_details}), achieving an $R^2$ value of 0.92 on the test set (10,000 mass maps); \cref{fig:modeling}C shows an example mass map.

We again construct an interpretable model using the wavelets distilled by AWD from the trained DNN.
To make predictions, we use the simple peak-counting algorithm developed in a previous work~\cite{ribli2019improved}, which convolves a peak-finding filter with the input images. Then, these peaks are used to regress on the outcome. 
In contrast to the fixed filters such as Laplace or Roberts cross used in previous works~\cite{ribli2019improved}, here we use the wavelets distilled by AWD, which result in three 2D wavelet filters (LL, LH, HL) and the 2D approximation filter (LL). The size of the distilled AWD filters is 12×12 and inspection of these filters shows a majority of the mass is concentrated on 3×3 subfilters (see \cref{fig:modeling}C). Then we extract those subfilters to use for peak-finding filters---by doing so, the size of the filters match with those used in~\cite{ribli2019improved} (additional details given in \cref{sec:cosmo_peak_count}). The hyperparameters for AWD are selected by evaluating the predictive model's performance on a held-out validation set.

\cref{tab:cosmo_results} shows the results of predicting using the peak-finding algorithm with various filters. The evaluation metric is the RMSE (Root mean square error). Its performance again outperforms the fully trained neural network (Resnet) model and the standard non-adaptive wavelet (DB5) model, as well as other baseline methods using Laplace filter and Roberts cross filter (see~\cref{sec:cosmo_peak_count} for details on how these filters are defined). Moreover, as can be seen in the compression rate, the AWD wavelet provides more efficient representations for the mass maps as well as concise explanation for the DNN's predictions compared to the DB 5 wavelet.

\begin{table}[H]
    \centering
    \caption{Performance comparisons for different models in cosmological parameter prediction. The lower RMSE and compression rate indicate better results. For RMSE, standard deviations are estimated from $10,000$ bootstrap samples.}
    
\begin{tabularx}{\textwidth}{lXXXXX}    
\toprule
& \textbf{AWD (Ours)} &  Roberts-Cross &   Laplace & DB5 Wavelet  & Resnet \\
\midrule
Regression (RMSE $\times 10^{-2}$)   &    \textbf{1.029 (0.033)} & 1.259 (0.039) & 1.369 (0.047) & 1.569 (0.048) &  1.156 (0.024) \\
Compression rate            &      \textbf{0.610}       &    N/A      &     N/A & 0.620 &  N/A \\
\bottomrule
\end{tabularx}
    
\label{tab:cosmo_results}
\end{table}
\cref{fig:modeling}C shows the learned AWD filters corresponding to the best distilled wavelet. The learned wavelet filters are symmetric and resemble the "matched filters" which have been used in the past to identify peaks on convergence maps in the cosmology literature~\cite{maturi2005optimal,schmidt2011weak}. We expect from cosmology knowledge that much information is contained in the peaks of the convergence maps (their amplitude, shape, and numbers), so this indeed matches our expectations based on physics. The high predictive performance further demonstrates that the AWD filters are more efficient at capturing cosmological information and better adapted to the shape of the peaks, than standard wavelets could do.

Moreover, the adaptive wavelet distillation allows us to look at "wavelet activation maps" (see~\cref{fig:cosmo_act_map}) to localize on locations in the convergence maps where important information is concentrated. In other words, we can indeed see that the AWD wavelet concentrates on identifying high intensity peaks, which is where most of the "localized" information is expected from theory.

\section{Discussion}
\label{sec:discussion}

In this work, we introduce AWD, a method to distill adapative wavelets from a pre-trained supervised model such as DNNs for interpretation. Doing so enables AWD to automatically detect and adapt to aspects of data that are important for prediction in an interpretable manner. The benefits of distilling relevant predictive information captured in a DNN are demonstrated through applications to synthetic and real data in two scientific settings. Overall, AWD allows us to interpret a DNN in terms of conventional wavelets, bringing interpretability with domain insights while simultaneously improving compression and computational costs, all while preserving or improving predictive power.

\paragraph{Future work} Here, we test our method with the saliency attribution method; many advanced interpretation techniques have been developed in the past years and the comparison between different interpretation techniques must be carefully explored in the context of a particular problem and audience. 
When optimizing the objective~\cref{eqn:objective} via gradient descent, it requires the gradient of the gradient, which is computationally expensive especially for large data and network sizes. A wavelet-based distillation approach that is computationally more amenable is an interesting direction for future research.
The current work learns a single-layer wavelet transform, but the complex nature of modern datasets often require strong nonlinearities.
Future work could extend AWD beyond a single-layer wavelet transform, e.g. by borrowing ideas from scattering transform~\cite{bruna2013invariant} or to other interpretable models~\cite{rudin2021interpretable,singh2021imodels}.
This would allow for bridging closer to deep learning while keeping interpretability, which can be effectively applied to other areas, such as computer vision and natural-image classification.
We hope to continue this line of research in order to improve the interpretability and computational efficiency of DNN models across many domains ranging from physical and biomedical sciences to computer vision and information technology.



\section{Acknowledgements}

We gratefully acknowledge partial support from NSF TRIPODS Grant 1740855, DMS-1613002, 1953191, 2015341, IIS 1741340, ONR grant N00014-17-1-2176. Moreover, this work is supported in part by the Center for Science of Information (CSoI), an NSF Science and Technology Center, under grant agreement CCF-0939370, and by the NSF Grant DMS 2031883 ``Collaboration on the Theoretical Foundations of Deep Learning''. SU was supported with funding from Philomathia Foundation and Chan Zuckerberg Initiative Imaging Scientist program. The authors would also like to thank Alan Dong for enlightening discussions on optimizing wavelets. The authors would also like to thank Tom Kirchhausen, Kangmin He, Eli Song, and Song Dang for providing the clathrin mediated endocytosis data to apply AWD for molecular partner predictions. 

\bibliographystyle{unsrt}
\bibliography{refs.bib}

\newpage




\appendix
\counterwithin{figure}{section}
\counterwithin{table}{section}
\begin{center}
	\Huge
	Appendix
\end{center}
\label{sec:supp}

\section{Further wavelet details}
\label{sec:wavelet_details_supp}
\renewcommand{\thefigure}{A\arabic{figure}}
\renewcommand{\thetable}{A\arabic{table}}

\subsection{2D wavelet transform} 
\label{sec:2d_wave_transform}

The discrete wavelet transform can be extended to two dimensions, using a separable (row-column) implementation of 1D wavelet transform along each axis. In 2D, the family of wavelets is characterized by the following three wavelets
\[\psi^1(x_1,x_2)=\phi(x_1)\psi(x_2), \ \psi^2(x_1,x_2)=\psi(x_1)\phi(x_2), \ \psi^3(x_1,x_2)=\psi(x_1)\psi(x_2), \]
named LH, HL, HH wavelets respectively. Together with the scaling function $\widetilde{\phi}(x)=\phi(x_1)\phi(x_2)$, the 2D discrete wavelet transform gives four components at each iteration, contrary to the 1D case, by applying the decomposition formula~\eqref{eqn:recursive_analysis} to the separable wavelets and scaling functions 
\begin{equation*}\label{eqn:recursive_analysis_2D}
	\begin{cases}
		a_{j+1}[p] = a_j \star \bar{h}\bar{h}[2p]; \\
		d^1_{j+1}[p] = a_j \star \bar{h}\bar{g}[2p]; \\
		d^2_{j+1}[p] = a_j \star \bar{g}\bar{h}[2p]; \\
		d^3_{j+1}[p] = a_j \star \bar{g}\bar{g}[2p],
	\end{cases}
\end{equation*}
for $p=(p_1,p_2)$, where for 2D discrete filters we denote ${h}{h}={h}[n_1]{h}[n_2]$. In particular, the decomposition yields three detail coefficients where the highpass filter $h$ is applied to either of the two-dimensional directions or both. These coefficients are intended to represent the signal in different orientations, i.e., vertical, horizontal, and diagonal. Similarly to~\eqref{eqn:recursive_synthesis}, the approximation coefficient $a_j$ at scale $2^j$ can also be recovered from the approximation coefficient $a_{j+1}$ and detail coefficients $d^k_{j+1}$, $k=1,2,3$, at scale $2^{j+1}$ with formula
\begin{equation*}\label{eqn:recursive_synthesis_2D}
	a_j[p] = [a_{j+1}]_{\uparrow 2}\star hh[p] + [d^1_{j+1}]_{\uparrow 2}\star hg[p] + [d^2_{j+1}]_{\uparrow 2}\star gh[p] + [d^3_{j+1}]_{\uparrow 2}\star g[p],
\end{equation*}
where $[a]_{\uparrow 2}$ denotes upsampling of the image $a$ by a factor $2$.

\subsection{Conditions for orthonormal wavelet basis}
\label{subsec:orthonormal_basis}

This section provides further details on constructing a valid wavelet $\psi$ such that the family $\{\psi_{j,n}\}_{(j,n)\in\Z^2}$ of wavelets forms an orthonormal basis of $L^2(\R)$. To do so, we introduce multiresolution analysis~\cite{mallat1989multiresolution,meyer1992wavelets} which constructs an orthonormal wavelet basis through approximations of signals at various resolutions. The key idea is that one builds a sequence of approximations for a signal with increasing resolutions while the difference between two consecutive approximations can be captured by the wavelet decomposition at a given scale.

To begin with, let $\phi$ be a scaling function in $L^2(\R)$. To motivate the idea of multiresolution analysis, we assume that $\phi$ is the Haar scaling function, defined as
\[\phi(t)=\begin{cases}
	1 & \text{if $0\leq t < 1$} \\
	0 & \text{otherwise}
\end{cases}. \]
Let $V_j$ denote the space spanned by $\{\phi_{j,n}\}_{n\in\Z}$, where $\phi_{j,n}(t)=2^{-j/2}\phi(2^{-j}t-n)$. Then $V_j$ is the set of piecewise constant functions over $[2^jn,2^j(n+1))$ for $n\in\Z$. The approximations of a signal $x$ at scale $2^j$ is defined by the orthogonal projection of $x$ on $V_j$, which is the closest piecewise constant function on intervals of size $2^j$. For two consecutive approximation spaces $V_j$ and $V_{j+1}$, the relation $V_{j+1}\subset V_j$ holds because any function that is constant over $[2^{j+1}n,2^{j+1}(n+1))$ is also constant over $[2^jn,2^j(n+1))$. Moreover, it is easy to see that $\lim_{j\to \infty}V_j=\{0\}$ and $\lim_{j\to -\infty}V_j=L^2(\R)$.

More generally, the sequence $\{V_j\}_{j\in\Z}$ of subspaces with $\{0\}\subset \ldots\subset V_{1}\subset V_0 \subset V_{-1}\subset \ldots \subset L^2(\R)$ is called a multiresolution approximation if it satisfies certain properties (see~\cite[Definition 7.1]{mallat2008wavelet}). The piecewise constant approximations induced by the Haar scaling function is a special case that verifies the properties of a multiresolution approximation. The multiresolution approximation is entirely characterized by the scaling function $\phi$ since the family $\{\phi_{j,n}\}_{n\in\Z}$ forms an orthonormal basis of $V_j$ for all $j\in\Z$. Remarkably, the following theorem due to~\cite{mallat1989multiresolution,meyer1992wavelets} further shows that a scaling function can be entirely determined by a discrete filter $h$ that is defined on the set of discrete values:

\begin{theorem}[Theorem 7.2\cite{mallat2008wavelet}]
	\label{thm:lowpass_sufficient}
	For a discrete filter $h[n]$, if the Fourier series $\widehat{h}(w)$ is $2\pi$ periodic and continuously differentiable in a neighborhood of $w=0$, if it satisfies
	\[\widehat{h}(0)=\sum_n h[n] = \sqrt{2} \ \ \text{ and } \ \ |\widehat{h}(w)|^{2}+|\widehat{h}(w+\pi)|^{2}=2 \ \text{ for all $w$}, \]
	and if $\inf_{w\in [-\pi/2,\pi/2]}|\widehat{h}(w)|>0$, then
	\[\widehat{\phi}(w)=\prod_{p=1}^{\infty}\frac{\widehat{h}(2^{-p}w)}{\sqrt{2}}, \]
	is the Fourier transform of a scaling function $\phi\in L^2(\R)$. Namely, the sequence $\{V_j\}_{j\in\Z}$ of subspaces induced by $\phi$ satisfies the properties of a multiresolution approximation.
\end{theorem}

Moreover, it can be shown that any scaling function $\phi$ determines the lowpass filter $h$ via $h[n] = \inner{\frac{1}{\sqrt{2}}\phi(t/2)}{\phi(t-n)}$ (see~\cref{eqn:filters}). Hence \cref{thm:lowpass_sufficient} provides equivalence
between the scaling function and the discrete lowpass filter. The next theorem states a necessary condition on the lowpass filter:

\begin{theorem}[Theorem 3\cite{burrus1997introduction}]
	\label{thm:lowpass_necessary}
	If $\phi$ is a valid scaling function, then
	\[\sum_{n} h[n] h[n-2k] = \begin{cases}1 & \text{if $k=0$} \\ 0  & \text{otherwise} \end{cases}.\]
\end{theorem}

\cref{thm:lowpass_sufficient} and \cref{thm:lowpass_necessary} characterize the sufficient and necessary conditions on the lowpass filter to build a valid scaling function.

Next, the multiresolution approximation requires $V_{j}\subset V_{j-1}$ for all $j\in\Z$ and the details that appear at the scale $2^{j-1}$ but disappear at the coarser scale $2^{j}$ can be characterized by the wavelet coefficients. Indeed, if $W_{j}$ denotes the orthogonal complement of $V_{j}$ in $V_{j-1}$, i.e., $V_{j-1}=V_{j}\oplus W_{j}$, one can construct a family of wavelets $\{\psi_{j,n}\}_{n\in\Z}$ that forms an orthonormal basis of $W_j$:

\begin{theorem}[Theorem 7.3\cite{mallat2008wavelet}]
	\label{thm:highpass_sufficient}
	Let $\phi$ be a scaling function and $h$ the corresponding filter. Let $\psi$ be the function having a Fourier transform
	\[\widehat{\psi}(w) = \frac{1}{\sqrt{2}}\widehat{g}\left(\frac{w}{2}\right)\widehat{\phi}\left(\frac{w}{2}\right),\]
	with 
	\[\widehat{g}(w)=e^{-iw}\widehat{h}^*(w+\pi).\]
	Then for any scale $2^j$, $\{\psi_{j,n}\}_{n\in\Z}$ is an orthonormal basis of $W_j$ and for all scales, $\{\psi_{j,n}\}_{(j,n)\in\Z^2}$ is an orthonormal basis of $L^2(\R)$.
\end{theorem}
In the time domain, the equation $\widehat{g}(w)=e^{-iw}\widehat{h}^*(w+\pi)$ can be converted to 
\begin{equation}
	\label{eqn:highpass_cond1}
	g[n]=(-1)^{n} h[N-1-n],
\end{equation}
where $N$ is the support size of $h$. Moreover, it follows from $\sum_n h[n]=\sqrt{2}$ (\cref{thm:lowpass_sufficient}) and $\sum_{n} h[n] h[n-2k]=\ones_{k=0}$ (\cref{thm:lowpass_necessary}) that $\sum_n h[2n]=\sum_n h[2n+1]$ holds~\cite[Theorem 2]{burrus1997introduction}. Using this identity, it is easy to check that the highpass filter must have zero-mean, i.e., 
\begin{equation}
	\label{eqn:highpass_cond2}
	\sum_n g[n]=0.
\end{equation}
Then \cref{eqn:highpass_cond1} and~\cref{eqn:highpass_cond2} provides the sufficient and necessary conditions on the highpass filter to build a valid wavelet $\psi$.

\section{Synthetic data details}
\label{sec:sim_details}
\renewcommand{\thefigure}{B\arabic{figure}}
\renewcommand{\thetable}{B\arabic{table}}

In this section, we show additional results for the experiments with synthetic data in~\cref{subsec:results_sim}. For this task, we generate data from a linear model $y_i=\inner{\Psi x_i}{\beta}+\epsilon_i$, $i=1,\ldots,n$, where:
\begin{itemize}
	\item The inputs $x_i\in\R^{n\times d}$ are generated with i.i.d. $\mathcal{N}(0,1)$ entries, where the number of input features is $d=64$;
	\item $\Psi$ is given a wavelet transform operator with DB 5 wavelets;
	\item The noise $\epsilon_i\in\R^n$ is generated with i.i.d. $\mathcal{N}(0,0.1^2)$ entries;
	\item The true coefficient $\beta$ is given $\beta_i=2$ for $3$ selected locations at a particular scale, and $\beta_i=0$ otherwise.
\end{itemize}
The data is randomly split into a training set of $50,000$ data points and a test set of $5,000$ data points. Then a $3$-layer fully connected neural network with $32$ hidden neurons each is trained on the training set with a learning rate of $0.01$ for $20$ epochs, achieving an $R^2$ score $>0.99$ on the test set. 

To distill the groundtruth wavelet (DB5) from this DNN, we solve the minimization problem given in~\cref{eqn:objective} for varying hyperparameters. Here we use a warm start strategy in which we solve the problem~\cref{eqn:objective} for one pair of values for hyperparameters $\lambda$ and $\gamma$ and use this solution to initialize the AWD filter at the next values of hyperparameters. In the initial stage of training, the AWD filter is initialized to the known lowpass filters corresponding to the DB 5 wavelet, Sym 5 wavelet, and Coif 2 wavelet, respectively (for DB 5, we add a noise to the lowpass filter). For each pair of the hyperparameters, the AWD filters were trained for $50$ epochs with Adam optimizer with a learning rate of $0.001$. All experiments were run on an AWS instance of p3.16xlarge for a few days.

\subsection{Additional results on synthetic data}
\label{sec:sim_wavelets}

\begin{figure}[H]
	\centering
	\includegraphics[width=\textwidth]{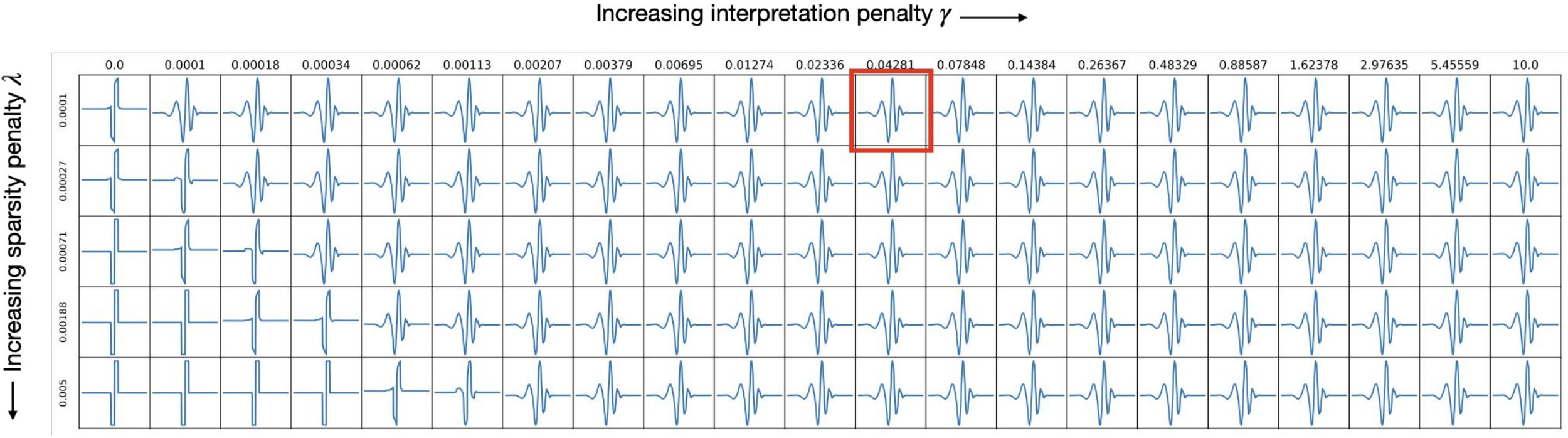}
	\caption{Varying sparsity and interpretation penalty yields different valid wavelets. In the initial stage, the AWD filter is initialized to the lowpass filter corresponding to DB 5 + noise.}
	\label{fig:sim_wave_vary_db5}
\end{figure}

\begin{figure}[H]
	\centering
	\includegraphics[width=\textwidth]{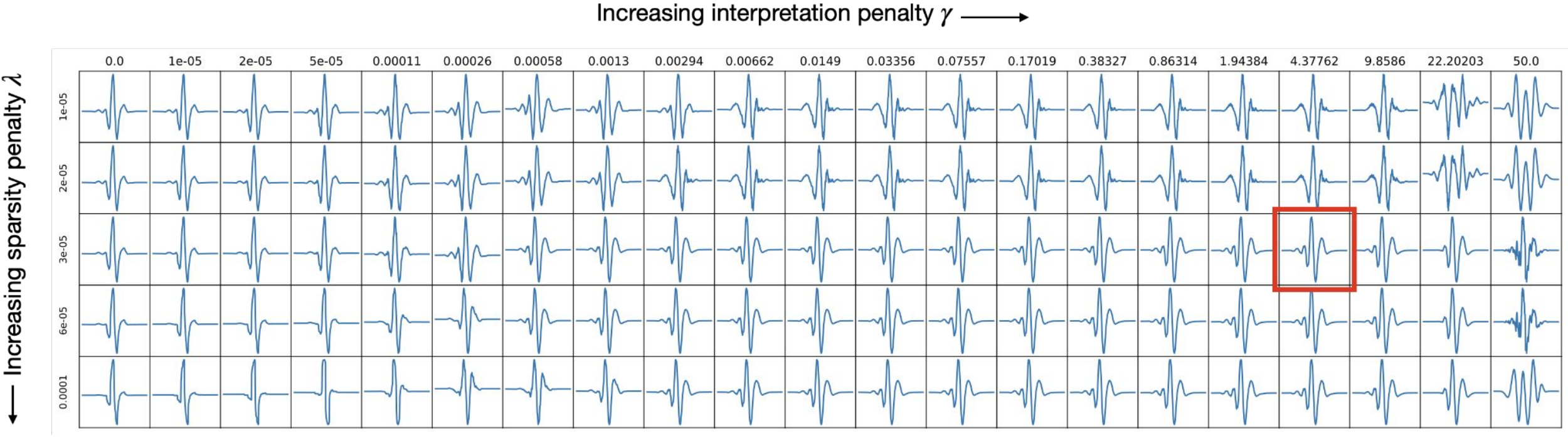}
	\caption{Varying sparsity and interpretation penalty yields different valid wavelets. In the initial stage, the AWD filter is initialized to the lowpass filter corresponding to Sym 5.}
	\label{fig:sim_wave_vary_sym5}
\end{figure}

\begin{figure}[H]
	\centering
	\includegraphics[width=\textwidth]{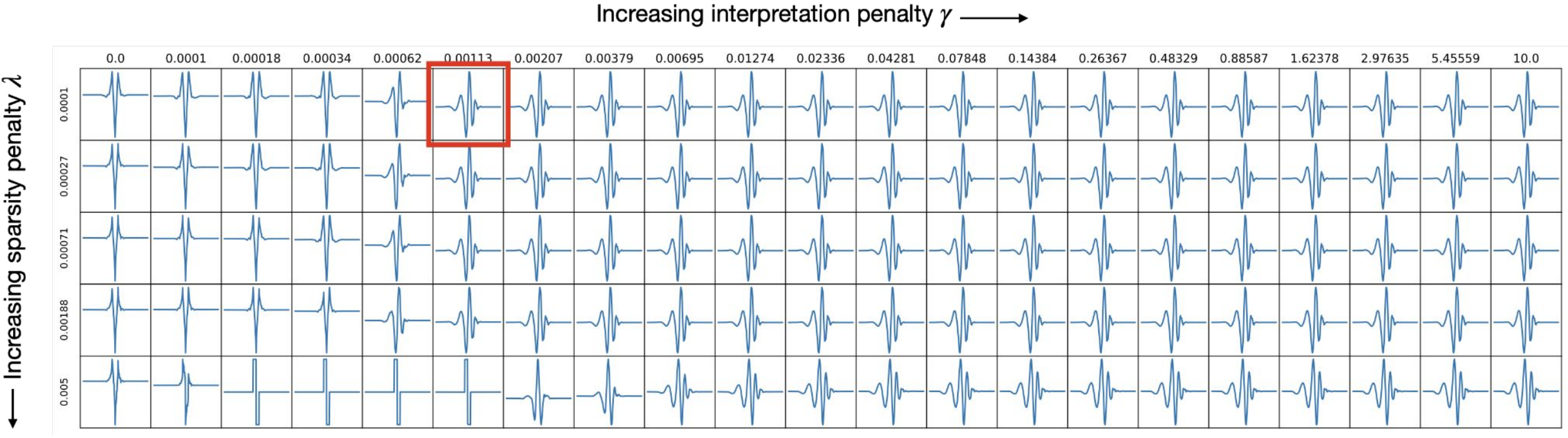}
	\caption{Varying sparsity and interpretation penalty yields different valid wavelets. In the initial stage, the AWD filter is initialized to the lowpass filter corresponding to Coif 2.}
	\label{fig:sim_wave_vary_coif2}
\end{figure}

Here we show the learned wavelets as the interpretation penalty $\gamma$ and the sparsity penalty $\lambda$ vary across a sequential grid of values spaced evenly on a log scale. \cref{fig:sim_wave_vary_db5} shows the results when the AWD filter in the initial stage is initialized to the lowpass filter corresponding to the DB 5 wavelet + noise; \cref{fig:sim_wave_vary_sym5} shows the results when the AWD filter in the initial stage is initialized to the lowpass filter corresponding to the Sym 5 wavelet; and \cref{fig:sim_wave_vary_coif2} shows the results when the AWD filter in the initial stage is initialized to the lowpass filter corresponding to the Coif 2 wavelet. We can see that as long as the interpretation penalty is not too small or large, the wavelets distilled by AWD accurately recovers the groundtruth (DB 5) wavelet. 

\begin{figure}[H]
	\centering
	\includegraphics[width=\textwidth]{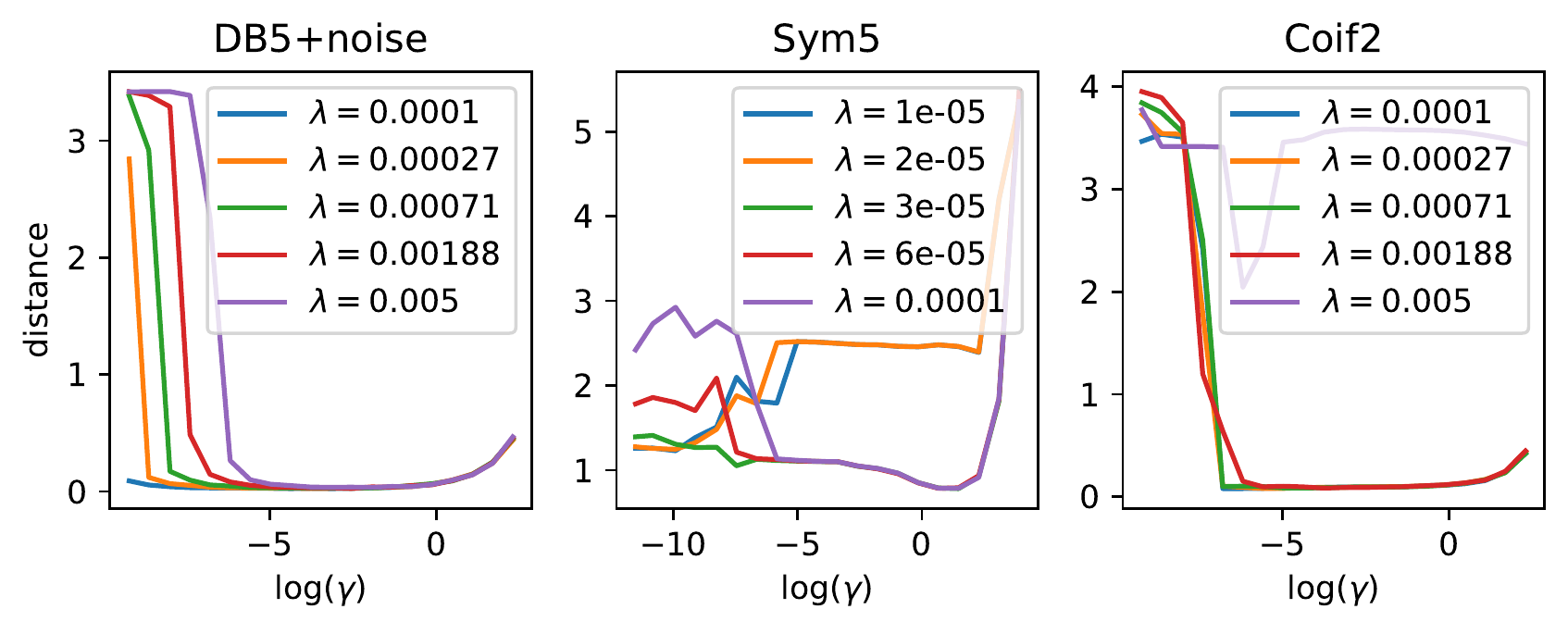}
	\caption{The distance between the learned wavelets and the groundtruth wavelet, defined as in~\cref{subsec:results_sim}, is plotted against $\log(\gamma)$ for different values of $\lambda$.}
	\label{fig:sim_dist}
\end{figure}

\cref{fig:sim_dist} calculates the distance between the learned wavelets and the groundtruth (DB5) wavelet, defined as in~\cref{subsec:results_sim}, as the interpretation penalty varies. When initialized at DB 5+noise, the learned wavelets get very close to the groundtruth wavelet for a wide range of $\gamma$ values, regardless of different sparsity penalty. On the other hand, when initialized at Sym 5, AWD can accurately recover the groundtruth wavelet only at the large values of $\lambda$; whereas for Coif 2, AWD can  recover the groundtruth wavelet only at the small values of $\lambda$.

\section{Molecular partner-prediction details}
\label{sec:bio_data_details}
\renewcommand{\thefigure}{C\arabic{figure}}
\renewcommand{\thetable}{C\arabic{table}}

\begin{figure}[t]
	\centering
	\begin{tabular}[b]{c}
		\includegraphics[width=0.45\textwidth]{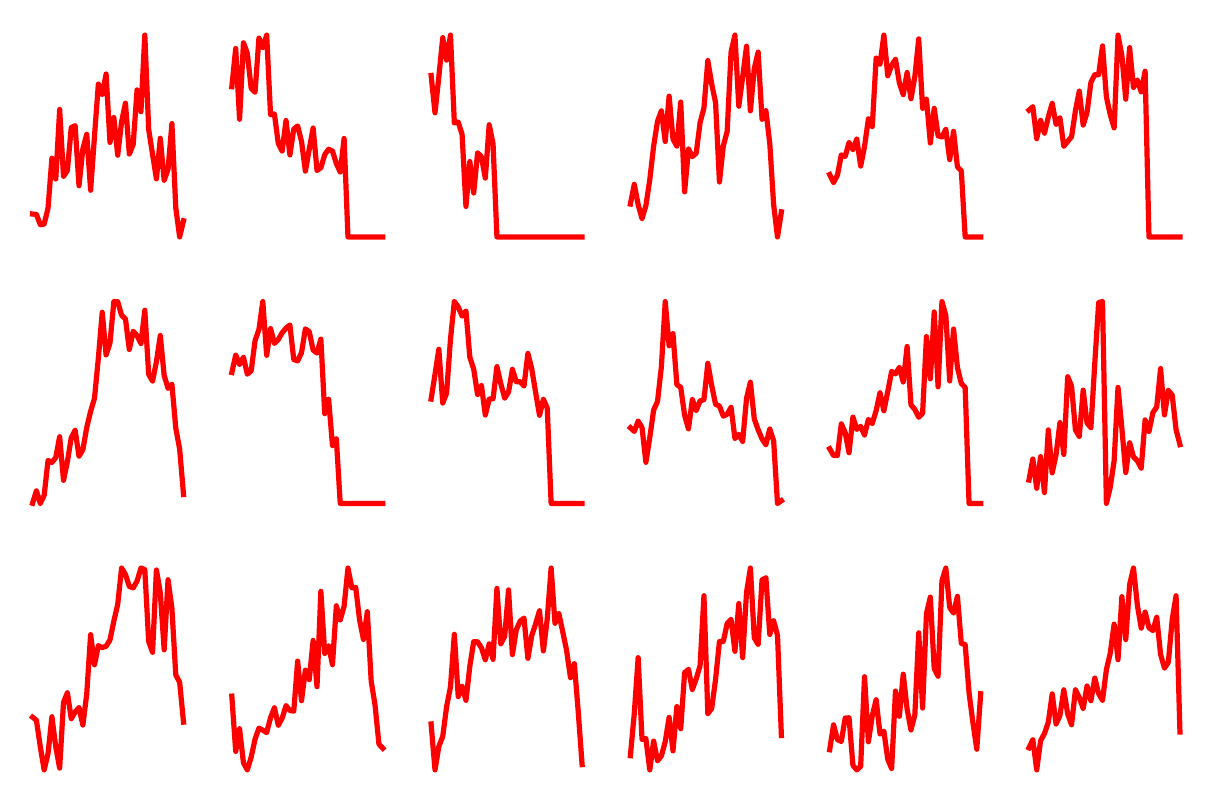} 
	\end{tabular}
	\caption{Fitted clathrin amplitudes for a few example events.}
	\label{fig:canonical_events}
\end{figure}

This section gives an overview of the preprocessing for the clathrin-mediated endocytosis problem in \cref{sec:bio_wavelets}.
For a detailed overview of the data, see the original study~\cite{he2020dynamics}.
In order to convert the raw fluorescence images to time-series traces, we use tracking code from previous work~\cite{aguet2013advances}. The tracking fits a Gaussian curve to the images (with standard deviation given by the imaging parameters). When the fit to the first channel (i.e. clathrin) is significant,\footnote{Here, significant is defined to be p-value less than 0.05, but the results are not sensitive to this precise threshold.} the track is recorded and a fit is forced to the second channel (i.e. auxilin). The amplitudes of each track over time are then extracted. \cref{fig:canonical_events} shows some examples of extracted clathrin traces.

The architecture of the LSTM used in this work has one recurrent layer, which takes an input of size 40 and has a hidden size of 40, followed by a single linear layer.

To train the AWD wavelet, the same warm start strategy was employed as in~\cref{sec:sim_details}. The AWD filters were trained for $100$ epochs with Adam optimizer with a learning rate of $0.001$. The experiment was run multiple times with respect to the randomness of mini-batches in the training procedure. All experiments were run on an AWS instance of p3.16xlarge for a few days.

\subsection{Distilled scaling functions and wavelets}
\label{sec:bio_best_wavelets}

Here we show the best wavelets selected by cross-validation and the corresponding scaling functions for $5$ different runs of the experiments. The results are stable across multiple runs, all capturing information about how rapid changes in the clathrin trace is useful for predicting the auxilin response.

\begin{figure}[t]
	\centering
	\includegraphics[width=0.4\textwidth]{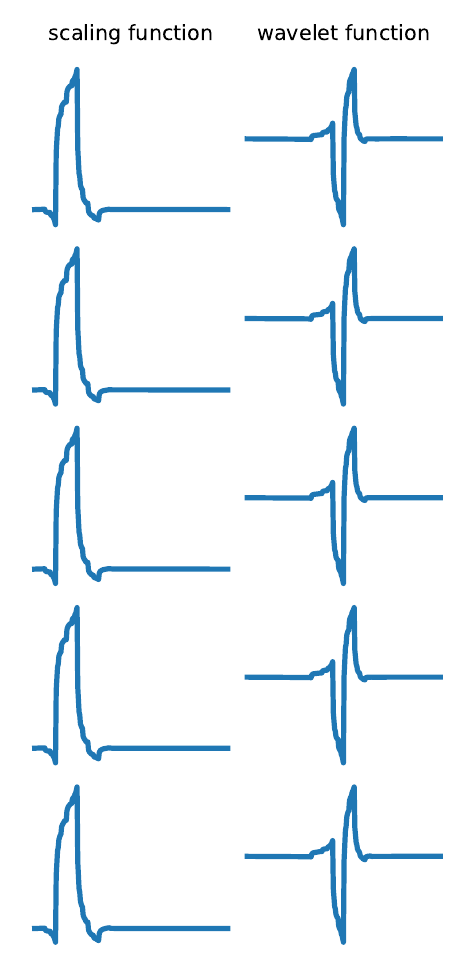}
	\caption{Optimal scaling and wavelet functions extracted by AWD across five random seeds.}
	\label{fig:bio_wave}
\end{figure}

\subsection{Varying sparsity and interpretation penalty}
\label{sec:bio_wavelets}

\begin{figure}[H]
	\centering
	\includegraphics[width=\textwidth]{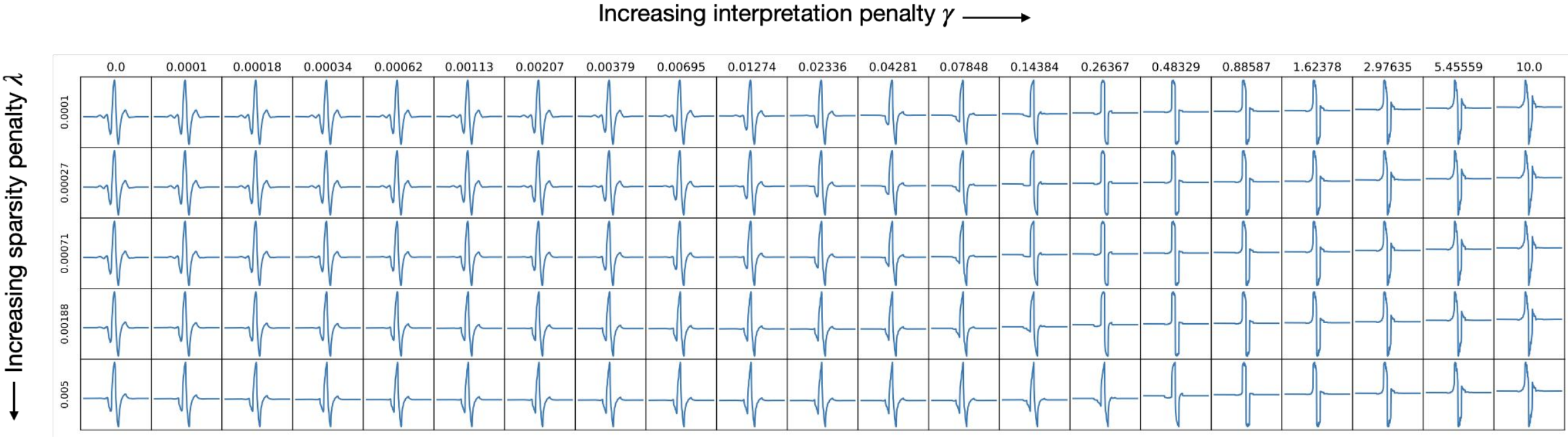}
	\caption{Varying sparsity and interpretation penalty yields different valid wavelets. In the initial stage of training, the lowpass filter is initialized to that corresponding to the Symlet 5 wavelet.}
	\label{fig:bio_wave_vary_sym5}
\end{figure}

\cref{fig:bio_wave_vary_sym5} shows the learned wavelets distilled by AWD as the interpretation penalty $\gamma$ and the sparsity penalty $\lambda$ vary. Unlike~\cref{fig:vary_wavelet} where the lowpass filter is initialized to the DB 5 wavelet in the initial stage of training, here the lowpass filter is initialized to that corresponding to the Sym 5 wavelet. For large values of $\gamma$, the learned wavelets captures qualitatively the same biological features as those shown in~\cref{fig:vary_wavelet}.

\subsection{Interpreting a single prediction}

\begin{figure}[H]
	\centering
	\includegraphics[width=0.91\textwidth]{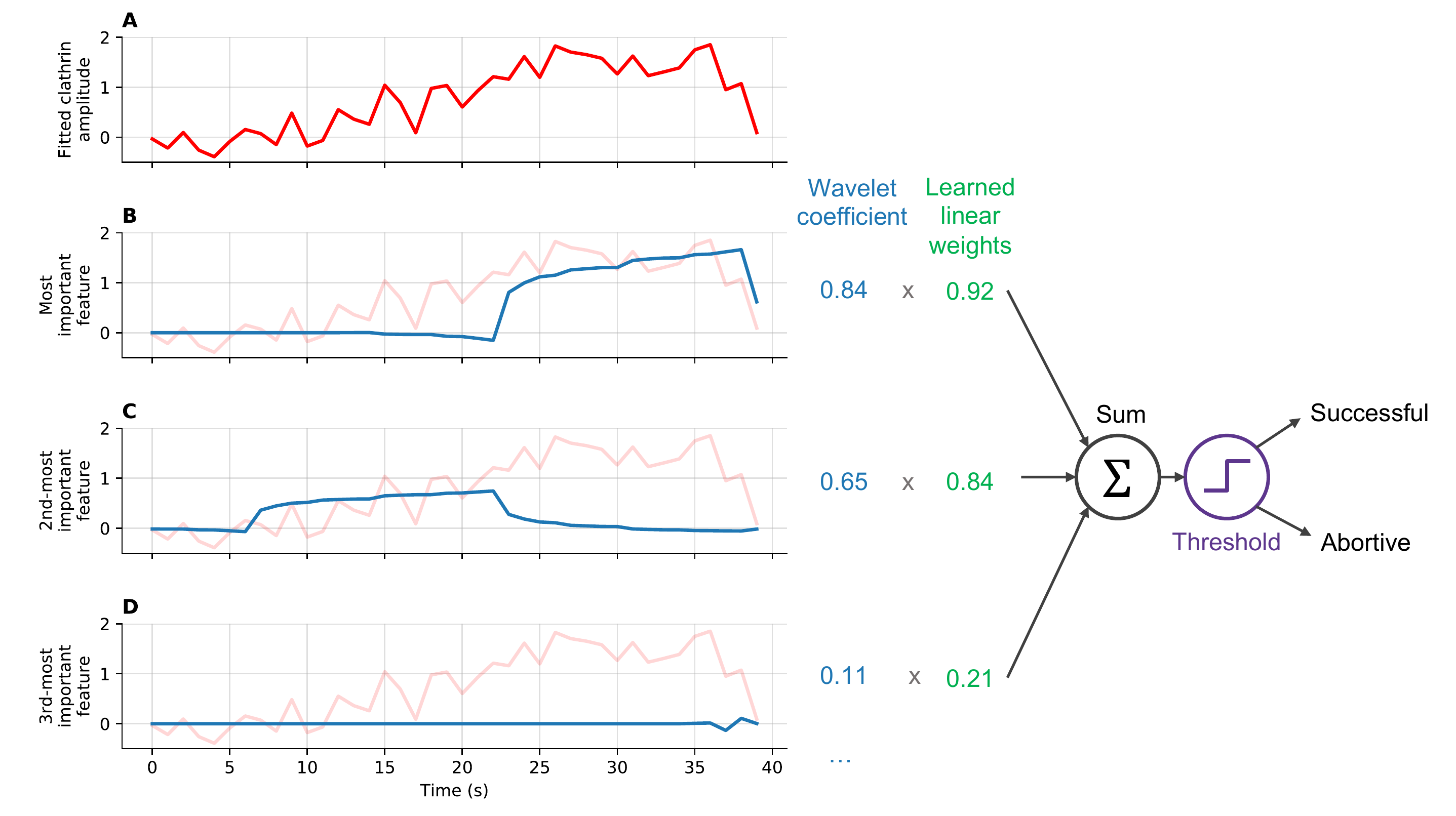}
	\caption{Interpreting a single prediction made by the wavelet model. The model takes the fitted clathrin amplitude shown in \bfp{A} and predicts that the event is successful. \bfp{B, C, D} show the three most important features for making this prediction.
		Each blue curve represents the input reconstruction for a single wavelet at a single scale. 
		The curves in \bfp{B} and \bfp{C} seem to capture meaningful components of the clathrin signal, as they find a gradual rise in the signal, a large peak in the signal, and finally a steep drop in the signal at the end.
		The model is simply a linear combination of wavelet coefficients: each blue curve yields a coefficient which is then multiplied by a learned weight. The final prediction of successful or abortive is then made by thresholding the sum of these products. 
		In this case, the first 2 coefficients dominate the prediction, and contributions for all remaining coefficients (some of which are omitted) are considerably less.
		For abortive predictions, the wavelet coefficients are usually much smaller (or negative).
	}
	\label{fig:my_label}
\end{figure}

\section{Cosmological simulation details}
\label{sec:cosmo_details}

\renewcommand{\thefigure}{D\arabic{figure}}
\renewcommand{\thetable}{D\arabic{table}}

For this task, we use the publicly available \texttt{MassiveNuS} simulation suite \cite{Jia2018}, composed of $101$ different $N$-body simulations spanning a range of cosmologies varying three parameters: the total neutrino mass $\Sigma m_\nu$, the normalization of the primordial power spectrum $A_s$, and the total matter density $\Omega_m$. These simulations are run at a single resolution of $1024^3$ particles for a $512$ Mpc/$h$ box size, and then ray-traced to obtain lensing convergence maps at source redshifts ranging from $z_s =1.0$ to $z_s=1100$.
To build our dataset, we select $10$ different cosmologies, listed in \autoref{tab:sim_params}, each of which provides $10,000$ mass maps at source redshift $z_s=1$. We rebin these maps to size $256\times 256$ with a pixel resolution of $0.8$ arcmin. 

\begin{table}[H]
	\centering
	
	\caption{Parameter values used in cosmology simulations.}
	
	\begin{tabular}{ccc}
		\toprule
		$m_\nu$ & $\Omega_m$ & $10^9A_s$\\
		\midrule
		0.0 & 0.3 & 2.1 \\
		0.06271 & 0.3815 & 2.2004 \\
		0.06522 & 0.2821 & 1.8826 \\
		0.06773 & 0.4159 & 1.6231 \\
		0.07024 & 0.2023 & 2.3075 \\
		0.07275 & 0.3283 & 2.2883 \\
		0.07526 & 0.3355 & 1.5659 \\
		0.07778 & 0.2597 & 2.4333 \\
		0.0803 & 0.2783 & 2.3824 \\
		0.08282 & 0.2758 & 1.8292 \\
		\bottomrule
	\end{tabular}
	\label{tab:sim_params}
\end{table}

For training the AWD wavelet, we use the same warm start strategy as in~\cref{sec:sim_details} while the initial lowpass filter is initialized to the lowpass filter corresponding to the DB 5 wavelet. The AWD filters were trained for $50$ epochs with Adam optimizer with a learning rate of $0.001$. All experiments were run on an AWS instance of p3.16xlarge for a few days.

\subsection{Peak counting algorithm}
\label{sec:cosmo_peak_count}

\begin{figure}[H]
	\centering
	\includegraphics[width=\textwidth]{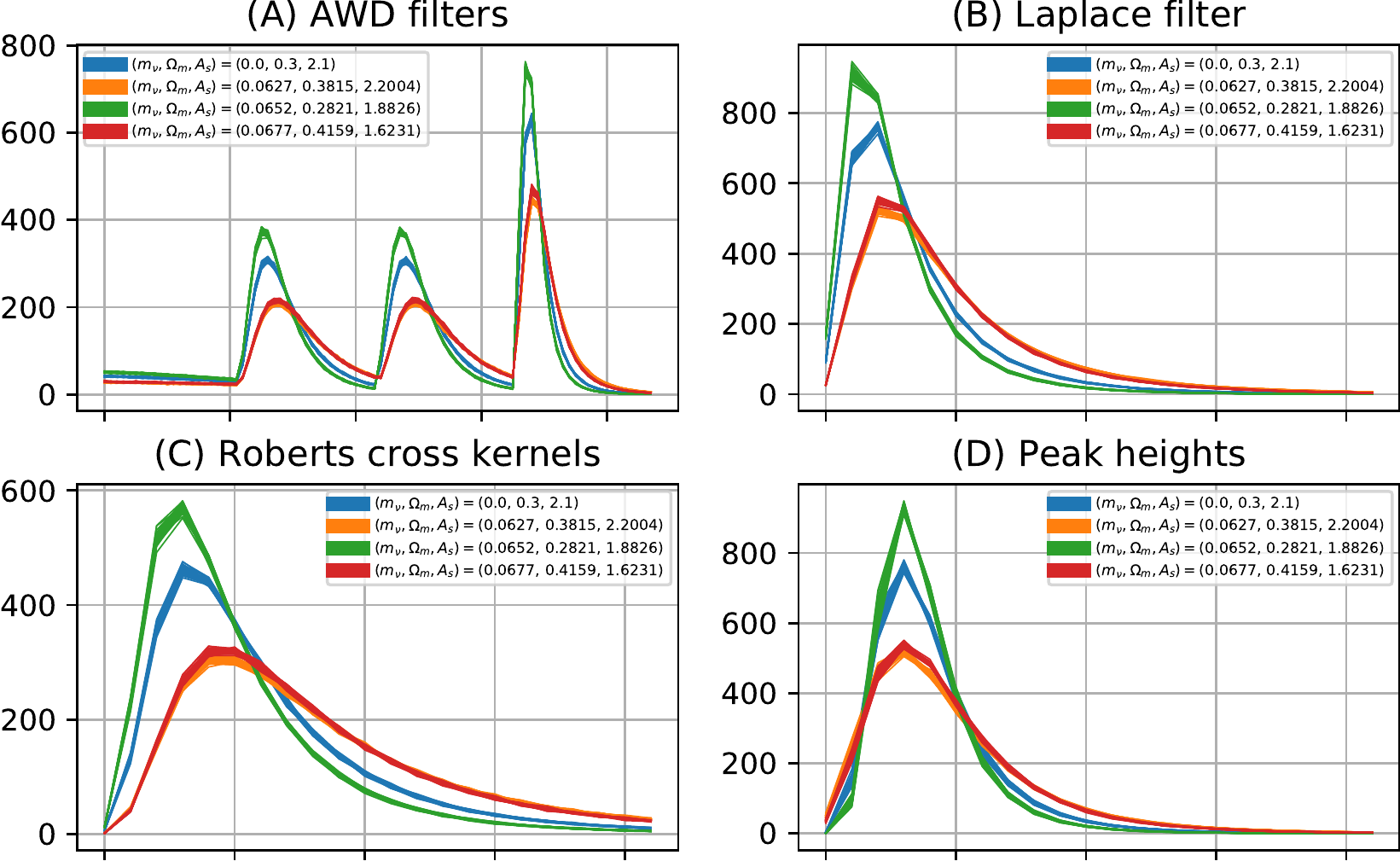}
	\caption{Peak steepness distributions using various filters.}
	\label{fig:cosmo_histograms_density}
\end{figure}

Here we describe the peak counting algorithm developed in~\cite{ribli2019improved} to compare the performance of various filters. In weak lensing, peaks are defined as local maxima on the lensing convergence maps. In the original peak counting algorithm, a histogram is made for each convergence map based on counting the raw pixel (height) values of the peaks on the maps (see~\cref{fig:cosmo_histograms_density}). At training time, the mean histograms and the covariance matrices are then created for each setting of the cosmological parameters $\xi=(m_\nu,\Omega_m,10^9A_s)$; and at test time, individual histograms are compared to the mean histograms via the distance 
\[d_{h,\xi}= (h-\mu_\xi)^\top \Sigma^{-1}_\xi (h-\mu_\xi), \]
and the parameters $\xi$ with the lowest distance $d_{h,\xi}$ is selected as prediction values. Here $h$ represents the histogram for a given map, and $\mu_\xi, \Sigma_{\xi}$, respectively, represent the mean histogram and the covariance matrix of the histograms for a cosmology with parameters $\xi$.

In~\cite{ribli2019improved}, the peak counting algorithm is generalized to exploit more information around the peaks compared with the height of the peaks. Inspired by the first layer of the trained CNN for parameter estimation, they propose to use peak steepness based on the isotropic Laplace filter, 
\[L=-\frac{10}{3}\left( \begin{array}{ccc} -0.05 & -0.2 & -0.05 \\ -0.2 & 1 & -0.2 \\ -0.05 & -0.2 & -0.05 \end{array}\right),\]
which computes the difference of the peaks and the surrounding pixel values, or the Roberts cross kernels,
\[R_x= \left( \begin{array}{cc} 0 & 1 \\ -1 & 0 \end{array}\right), R_y= \left( \begin{array}{cc} 1 & 0 \\ 0 & -1 \end{array}\right),\]
which compute the gradient at the peaks. For the Laplace filter, the peak steepness values are calculated via convolving the filter with the input images at the position of the peaks. For the Roberts cross kernels, the two filters $R_x$ and $R_y$ are applied to the $4$ adjacent $2\times 2$ pixel blocks around the peaks and the magnitudes are calculated via $G_i=\sqrt{G_{x,i}^2 + G_{y,i}^2}, i=1,\ldots,4$, where $G_{x,i}$ and $G_{y,i}$ are the sub-images after convolve $R_x$ and $R_y$ with the $i$-th adjacent pixel blocks. Then the sum of the $4$ magnitudes $\sum_{i=1}^{4}G_i$ is used to get the peak steepness values. 

Here we further use the wavelet filters distilled by AWD as peak-finding filters in the peak counting algorithm. To match the size of the distilled AWD filters with that of the Laplace filter or Roberts cross kernels, we extract $3\times 3$ subfilters from the wavelet filters where a majority of the mass is concentrated on. This results in $4$ different $3\times 3$ filters, corresponding to three wavelet filters (LH,HL,HH) and one approximation filter (LL), which are then used as peak-finding filters to calculate the histograms of the peak steepness values. \cref{fig:cosmo_histograms_density} shows the distributions of peak steepness values using various filters mentioned above.

To run the peak-counting algorithm with various filters, we need to select the number, width, and range of bins. For the Laplace filter and Roberts cross kernels, we use the same settings as~\cite{ribli2019improved} which runs bins from $0$ to $0.22$ in $0.01$ wide. In the case of the wavelet filters, we keep the same number of bins while the range is chosen via the algorithm’s performance on a held-out validation set. The resulting bin is then used to evaluate the prediction performance on the test set.  

\subsection{Wavelet activation maps}
\label{sec:cosmo_act_maps}

\begin{figure}[H]
	\centering
	\includegraphics[width=0.33\textwidth]{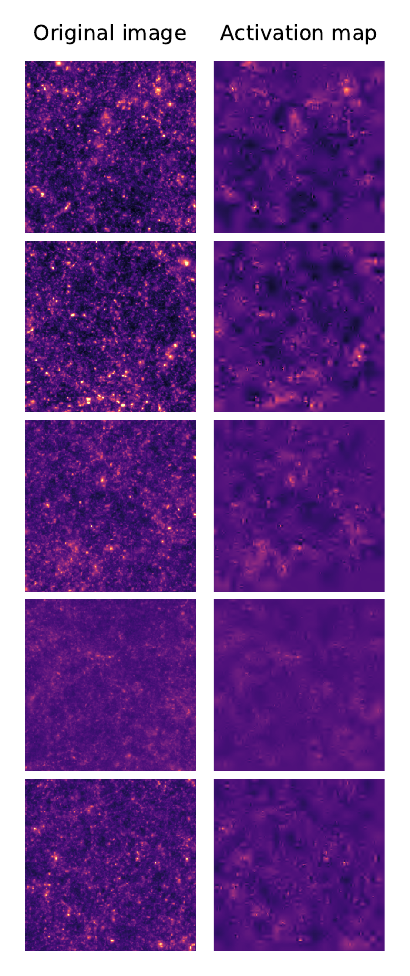}
	\caption{Wavelet activation maps for individual images made by the AWD model.}
	\label{fig:cosmo_act_map}
\end{figure}

As part of our interpretability analysis, we now show images that highlight important features for predicting $\Omega_m$ (total fraction of matter in the universe) in~\cref{fig:cosmo_act_map}. To create the images, for each map we calculate feature attributions on the wavelet domain extracted by AWD using TRIM (here we use IG~\cite{sundararajan2016gradients} to get attributions). Then only the wavelet coefficients with top $600$ attributions (out of $73,839$) are retained to transform back to the image domain using inverse wavelet transform. We can see that the activation maps highlight localized regions in the original maps that correspond to the high intensity peaks and voids. This is consistent with the known cosmology theory that these peaks contain high constraining power to predict cosmological parameters of the universe.

\end{document}